\documentclass{article}

\usepackage{arxiv}

\usepackage[utf8]{inputenc} % allow utf-8 input
\usepackage[T1]{fontenc}    % use 8-bit T1 fonts
\usepackage{hyperref}       % hyperlinks
\usepackage{url}            % simple URL typesetting
\usepackage{booktabs}       % professional-quality tables
\usepackage{amsfonts}       % blackboard math symbols
\usepackage{nicefrac}       % compact symbols for 1/2, etc.
\usepackage{microtype}      % microtypography
\usepackage{lipsum}
\usepackage{graphicx}
\usepackage{subcaption}
\usepackage{xcolor}
\usepackage{multirow}
\graphicspath{ {./images/} }

\title{QBERT: Generalist Model for Processing Questions}

\author{
     Zhaozhen Xu \\
  Intelligent Systems Laboratory\\
  University of Bristol\\
  Bristol, BS8 1QU, UK\\
  \texttt{zhaozhen.xu@bristol.ac.uk} \\
   \And
 Nello Cristianini \\
  Intelligent Systems Laboratory\\
  University of Bristol\\
  Bristol, BS8 1QU, UK\\
  \texttt{nello.cristianini@bristol.ac.uk} \\
}

\begin{document}
\maketitle
\begin{abstract}
Using a single model across various tasks is beneficial for training and applying deep neural sequence models. We address the problem of developing generalist representations of text that can be used to perform a range of different tasks rather than being specialised to a single application. We focus on processing short questions and developing an embedding for these questions that is useful on a diverse set of problems, such as question topic classification, equivalent question recognition, and question answering. This paper introduces QBERT, a generalist model for processing questions. With QBERT, we demonstrate how we can train a multi-task network that performs all question-related tasks and has achieved similar performance compared to its corresponding single-task models.
\end{abstract}

% keywords can be removed

\section{Introduction}
There is increased attention to the problem of learning generalist agents (as opposite to specialist) in a way that the same representation can be used in a range of tasks, even if it does not excel at any specific task \cite{reed2022generalist}. While a specialist should be expected to excel at its one task, a generalist is expected to be good at many problems. In this paper, we focus on building a generalist model for processing a special type of short text: Question.

The development of online communities produces a massive amount of text every day. For example, in the question domain, with the rise of commercial voice assistants such as Siri and Alexa and communities such as Quora, numerous questions are asked on a daily basis. Processing these questions can provide a new perspective on understanding communities and people's interests.

In this paper, we define the generalist model as a question-processing model that targets analysing the semantic and syntactic information in the question. More specifically, this generalist model can process the questions in terms of question topic classification, equivalent question recognition, and question answering, which will be explained in section \ref{task}.

Some state-of-the-art deep learning models like Transformer \cite{vaswani2017attention} are widely used in natural language processing (NLP). They resulted in leading performance for various tasks \cite{radford2018improving,devlin2018bert,yang2019xlnet}. Train a language model requires lots of training data. Therefore, researchers had to create pre-trained language models using large-scale unsupervised tasks and then fine-tune them with labelled task-specific data. However, labelled data for a specific task are always limited and hard to obtain. Besides, a language model can have a size of millions or billions of parameters. It is usually expensive to train and use a separate network for each task. A generalist model can help address these problems by applying multi-task learning, a learning approach that improves generalisation by adding inductive bias such as tasks and domain information \cite{caruana1997multitask}.

There are two main strategies for multi-task learning. One standard approach is adding extra tasks, also referred to as auxiliary tasks, to improve the performance of the target task. Empirically, adding auxiliary tasks to a pre-trained network is more similar to transfer learning, which improves primary tasks with additional tasks. Another \cite{mccann2018natural} is learning all the tasks jointly without identifying the primary task so that all the tasks can achieve balanced performance, which can be leveraged for training a generalist agent.

We fine-tune the pre-trained language model with all the tasks jointly without identifying primary and auxiliary tasks. These tasks share the same domain, which is referred to as inductive bias multi-task learning \cite{redko2019advances}. Research \cite{liu2019multi,raffel2020exploring} shows that multi-task learning and pre-trained language models are complementary and can be combined to generate better performance on learning text representations. 

There are many different types of tasks included in multi-task natural language understanding. For example, single sentence classification like sentiment analysis, pairwise classification like natural language inference, and regression task like sentence similarity. MT-DNN \cite{liu2019multi} trains their multi-tasking model with the transformer encoder and task-specific layer so that it can apply to classification and regression tasks. To adapt to various tasks,  some researchers re-frame all the datasets into the same format. MQAN \cite{mccann2018natural} formulates all the datasets into question answering over context. T5 \cite{raffel2020exploring} creates a sequence-to-sequence format for the tasks. All these models focus on general language understanding tasks like GLUE \cite{wang2018glue}, and decaNLP \cite{mccann2018natural}. 

In contrast, we focus on a range of different tasks for processing questions. And we report here on a generalist network called QBERT to solve three processing tasks we defined in the question domain. QBERT intends to work as a ``generalist'' language model that can perform multiple question tasks rather than a ``specialist'' who is only trained to maximise the performance on one specific task.

QBERT is based on sentence-BERT (SBERT) \cite{reimers2019sentence}, a Siamese BERT (Bidirectional Encoder Representations from Transformers) \cite{devlin2018bert} that projects the sentences into high-dimensional vector space. This process is known as embedding. The sentence embeddings with similar semantic meanings are close to each other in the vector space. Note that our intention is not to design a new algorithm but to fine-tune SBERT in a multi-task way so that the same representations can be used for processing questions in multiple ways. After fine-tuning SBERT, the embeddings generated from the input sequence can be used for both classification and retrieval tasks. 

A previous study \cite{xu2021makes} on the question-related multi-tasking model shows that the training curriculum is critical. They reported that one certain curriculum could obtain a balanced performance on all the tasks. However, one of the limitations of the previous study is that the model lacked consistency on different question tasks. Reference \cite{xu2021makes} performed topic classification with a single BERT structure, others with Siamese BERT. To improve this, we re-frame the single sentence classification into a retrieval task.

During inference, QBERT produces the representation of the input sequence without any task-specific modules. Instead, it contains a threshold filter to determine the cosine similarity of the embedding pairs. Compared to the standard multi-task structure, reducing task-specific layers simplifies the complexity of the network. The network shares all the weights between tasks, also known as hard parameter sharing. More details of QBERT will be explained in section \ref{QBERT}.

After that, we compare QBERT with SBERT and the single-task version of SBERT in section \ref{result}. The results in section \ref{performance} also show how the training curriculum affects the performance of QBERT.

\section{Tasks}\label{task}

In this paper, we define task (T) by data (X), label (Y), and loss function (L) as follow.

\begin{equation}
    T\doteq \{ p(X), p(Y\mid X), L\}
\end{equation}
Where $p(X)$ is the distribution of the input data, $p(Y\mid X)$ is the distribution of label $Y$ given data $X$, and $L$ is the loss function.

QBERT combines 3 different types of tasks: \textbf{question topic classification}, \textbf{equivalent question recognition}, and \textbf{question answering}. These tasks target common natural language understanding problems such as single sentence classification, pairwise classification, and information retrieval.

{\bfseries Question Topic Classification (QT):} Given a question, the model labels the topic of the question. 
% The questions are categorised into 10 different topics according to \textit{Yahoo! Answer} \cite{zhang2015character}.

{\bfseries Equivalent Question Recognition (QE):} There are two sub-tasks included in QE, classification and retrieval. In classification, the model aims at classifying if the question pairs are similar or not, and based on the outcome, retrieve all similar questions from a question corpus with the given question. 
% The model is expected to group similar questions in the vector space.

{\bfseries Question Answering (QA):} Given a question, the model searches for the answer from lists of candidate sentences. We determine this task as an open-domain open-book QA in which the question has no limitation in domains; the model allows answering the question with the content provided. 
% For retrieval, the model picks one sentence from the given context as the answer.
% The QA task includes both retrieval and classification as well. 

\section{QBERT: a multi-task question-processing version of BERT}\label{QBERT}

Our model is inspired by SBERT \cite{reimers2019sentence}, which projects input sequence (sentence in this case) in high dimensional space. In such a way, we can evaluate the similarity between input sequences in the vector space using cosine similarity and retrieve the most similar sequence within a given corpus. Additionally, we can apply our model in classification by introducing a similarity threshold. 

The three question tasks we defined in the previous section include three kinds of machine learning tasks: single sentence multiclass classification, pairwise classification, and information retrieval. In the previous research \cite{xu2021makes}, topic classification was performed with a separate network because it is a multiclass classification that requires single input instead of pairwise. To perform these three tasks with one Siamese model, we consider the topic classification as pairwise classification by taking the (Question, Topic) as the pairwise input. During inference, instead of categorising the topic of a question, we retrieve the closest topic to a question.

\begin{figure}[!h]
    \centering
    \includegraphics[width = 0.99\textwidth]{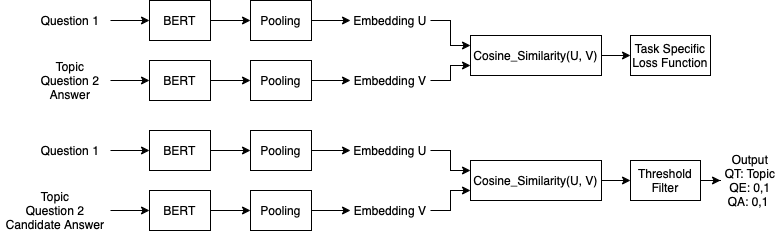}
    \caption{QBERT architecture. The architecture is based on SBERT but trained to have a balanced performance on various tasks.Top: Training as binary classification, Bottom: Inference by calculating the cosine similarity between input sequences. All BERTs share the same parameters.}
    \label{fig:qbert}
\end{figure}

% The model is constructed with both shared layers among different tasks and task-specific layers for different loss functions during training. 
% Furthermore, in conjunction with approximate nearest neighbour search methods \cite{JDH17}, QBERT is time-efficient in retrieving answers from large corpus such as Wikipedia.
Figure \ref{fig:qbert} illustrates the architecture of QBERT. During training, all the tasks are trained as to minimise the cosine distance using the binary labels and update the shared BERT layer. Task-specific loss functions are introduced for different types of data. While inference, the model only requires shared layers without task-specific layers, which manages to simplify the model.

{\bfseries Input layer:} $S = ( s_{1}, ..., s_{n})$ is an input sequence with n words. The sequence can be either a topic, question, sentence, or paragraph. The model takes a pairwise input $(S, S')$ such as a question pair, question-topic pair, or question-answer pair. The pairwise input is then passed to two identical BERTs.
% According to BERT, the $s_{1}$ token is always a [CLS] classification token. The word tokens include word embedding, position embedding, and segment embedding. 

{\bfseries BERT layer:} BERTs in this layer share all the parameters. The shared embedding layer following the setup of $BERT_{base}$ which takes the sequence input as word tokens and generates an output for each token as well as a [CLS] token at the beginning of the output sequence. $BERT_{base}$ uses the an encoder containing 12 layers and 110M parameters and is pre-trained with two unsupervised tasks: masked language model and next sentence prediction. The output of the BERT layer is in $\mathbb{R}^{d}$ vector space, and according to BERT, $d = 768$.

{\bfseries Pooling layer:} Similar to SBERT, the model leverages a mean pooling strategy that computes the mean of all output tokens (except [CLS]) of the sequence from BERT. According to SBERT, the mean pooling strategy outperforms using the [CLS] token as the embedding on capture sequence similarity. After the pooling function, the model generates a pair of embedding U and embedding V as equation \ref{eqn:embed} , where $U\in \mathbb{R}^{d}$ and $V\in \mathbb{R}^{d}$.
\begin{equation}
\label{eqn:embed}
Embedding = \frac{1}{n}\sum_{i=1}^{n} \Phi_{BERT}(s_{i})
\end{equation}

We apply two different loss functions for different types of data: online contrastive loss for binary classification tasks that have both positive and negative sample, and multiple negatives ranking loss for information retrieving datasets that does not contain positive nor negative label. Adam optimiser \cite{kingma2014adam} minimises the loss based on the cosine similarity $D_{cosine}(U,V)$.
% \begin{equation}
%     D_{cosine}(U,V) = \frac{\sum_{i =1}^{n}U_{i}V_{i}}{\sqrt{\sum_{i=1}^{n}U{_{i}}^{2}}\sqrt{\sum_{i=1}^{n}V{_{i}}^{2}}}
% \end{equation}

{\bfseries Pairwise classification specific layer:} QBERT introduces the contrastive loss \cite{hadsell2006dimensionality} for pairwise classification. It aims to gather positive pairs in the vector space while separating negative pairs. For embedding U, V, the loss is calculated as follows.
\begin{equation}
\label{eqn:l1}
    L_{contrastive}=\frac{1}{2}\left \{ Y\left ( 1-D_{cosine} \right )^{2} + \left ( 1-Y \right )\left [ max\left ( 0, m-(1-D_{cosine}) \right )\right ]^{2}\right\}
\end{equation}
Where $Y$ is the binary label. $Y = 1$ if $U$ and $V$ are related. And the distance $D = 1-D_{cosine}$ between $U$, $V$ is minimised. When $Y = 0$, the distance increases between $U$, $V$ until larger the given margin $m$. In particular, we apply online contrastive loss that only computes the loss between hard positive and hard negative pairs.

{\bfseries Retrieval specific layer:} One of the advantages of applying multiple negative ranking loss is that the training dataset no longer requires either positive or negative labels. For a given positive sequence pair $(S_{i}, S_{i}')$, the function assumes that any $(S_{i}, S_{j}')$ is negative when $i \neq j$. For example, in QA, for question set $Q = \{q_{1}, ..., q_{m}\}$ and answer set $A = \{a_{1}, ..., a_{m}\}$, $(q_{i}, a_{i})$ is a positive pair given by the dataset, $(q_{i}, a_{j})$ is a negative pair randomly generated from the dataset. The cross-entropy loss of all the sequences pairs is calculated as follows.
\begin{equation}
\label{eqn:l2}
    L_{multiple\_negative}=-(Ylog\left ( D_{cosine} \right ) + \left ( 1-Y \right ) log(1-D_{cosine}))
\end{equation}
During inference, QBERT no longer leverages a task specific layer. Instead, it introduces a threshold filter. QBERT calculates the $D_{cosine}(U, V)$, the cosine similarity between embeddings $U$ and $V$, and applies different similarity thresholds for each task to determine if two sequences are related in terms of topic, equivalent question, or corresponding answer. The threshold of best performance is selected after training.

\subsection{Training Curriculum}
During training, the data in each dataset get divided into batches $B = \{b_{1}, ..., b_{n}\}$. In each step, one batch $b_{i}$ is selected randomly, and the model parameters are updated by stochastic gradient descent. As shown in the previous research \cite{xu2021makes}, the training curriculum was critical for multi-task question processing. In reference \cite{xu2021makes}, the tasks were trained once at a time, from QE to QA to QT (QT was trained with different network architecture). However, the tasks learned in the earlier stage had a worse performance compared to the tasks learned in the later stage. To improve this, we train QBERT in a fixed-order round robin (RR) curriculum and compare the results with one by one (OBO) curriculum.

In the OBO approach, we train QBERT following QT, QE, and QA orders. Every dataset is divided into multiple batches, each with specific batch size, and is trained one at a time. The parameters are updated during the training and shared amongst all the datasets.

On the contrary, QBERT trains all the tasks simultaneously in the RR curriculum. The data in each task-specific layer are built as mini-batches and divided into two task-specific layers. During each step, the model is trained and updated by batches from both classification and retrieval tasks. QBERT-RR alternates between tasks during training which prevents the model from forgetting about the tasks learned at the beginning of the training.

\subsection{Threshold Filter}
\label{threhsold}
% After training the model, we define the thresholds used for inference in QE and QA, respectively. 

In QE and QA, apart from classifying if the sequences are related, it is also crucial for the system to search all the related sequences (equivalent question or answer) for the given questions. The problem is how to quantify ``related'' with embeddings. A cosine similarity threshold is introduced in this model. Using a threshold simplifies the network structure of QBERT during inference by removing the task-specific layer. The threshold filter acts like a margin separating related $(S, S')$ from others. Furthermore, with this threshold, the network can not only search the information that is closest to the query but can also identify if the information is related (close enough) to the query. For example, a question might be unique in the corpus so that the closest question to the given question is not equivalent to the given question if it has a smaller cosine similarity than the threshold; or a question might not have a high-confidence answer from the candidate corpus, the closest candidate with a cosine similarity smaller than the threshold will not be considered as the right answer.
% As illustrated in figure \ref{fig:qbert}, the input pairs share an identical BERT among different tasks.

To decide the threshold, first, all the sequences in the training set are embedded with the fine-tuned model. The sequence pairs are classified as positive if they have more similarity than the threshold. The similarity threshold with the best accuracy in the training set is found to quantify any question pairs during testing. With a threshold, the model is capable of searching and grouping all the related sequences in a given candidate corpus.

\section{Experiments}

\label{result}

Training QBERT includes pre-training and multi-task training. We follow the pre-training of BERT and SBERT. Then we perform multi-task learning on five question related datasets and evaluate on four of them.

\subsection{Datasets}
% This section describes the datasets we use to train QBERT. 
{\bfseries Quora Question Pair (QQP) \cite{quora}} first released on Quora in 2017. It is a dataset that contains 404k question pairs collected and annotated by Quora. QQP labels if the questions are duplicated or not. There are 537k unique questions in the dataset. Training on QQP, we aim at improving the performance of QE tasks for QBERT. We then evaluate QQP in both pairwise classification and equivalent questions retrieval.

{\bfseries WikiQA \cite{yang2015wikiqa}} is a question-answering dataset which has the questions from query logs on Bing and answers from Wikipedia's summaries. The questions in WikiQA are factual questions that start with WH words like who, what, and when etc. The candidate answers are extracted sentences from the first paragraph of Wikipedia articles (also known as Wikipedia Summary). The dataset includes 3,047 questions and 26k candidate sentences, of which 1,239 questions contain a correct answer. We train WikiQA as a classification task and evaluate it as an answer selection task.

{\bfseries Yahoo! Answer \cite{zhang2015character}} data were originally collected by Yahoo! Research Alliance Webscope program. Zhang et al., built up a corpus which contains 1.46M samples within 10 most popular topics on \textit{Yahoo! Answer}. The sample includes the topic, question title, question content, and the best answer provided by the user. We apply this corpus in both QT and QA tasks. For QT, QBERT takes the question title and topic as the input sequence pairs. Question title and best answer are leveraged for training QA tasks. To distinguish the data used in different tasks, we use YT for the data applied in QT and YQA for data in QA.

{\bfseries Stanford Question Answering Dataset (SQuAD) \cite{rajpurkar2016squad}} is a corpus that contains questions, answers and contexts for reading comprehension tasks. The contexts are extracted from Wikipedia. We use SQuAD 1.1, which all the questions have a corresponding answer phrase in the given context. There are 98,169 question-answer pairs in the dataset. To train QBERT, we take the question and the one sentence in the context that contains the answer phase as input.

\begin{table}[hbt]
\caption{Statistics of the training datasets. Note that the test set of SQuAD is confidential from researcher. The number of test data states here is the validation set that is publicly accessed. The metrics for SQuAD in this paper is ``exact match in sentence'' which will defined in section \ref{performance}.}
\centering
\label{tab:dataset}
\begin{tabular}{|l|l|l|l|l|}
\hline
\textbf{Dataset} & \textbf{\#Train}    & \textbf{\#Test}  & \textbf{Label} & \textbf{Metrics}      \\ \hline
YT      & 1,400,000 & 60,000 & 10    & Accuracy    \\ \hline
QQP     & 283,001    & 121,286 & 2     & Accuracy/F1   \\ \hline
WikiQA  & 23,080     & 6,116   & 2     & Accuracy/F1   \\ \hline
YQA     & 14,000,000 & 600,000 & 1     & -           \\ \hline
SQuAD   & 87,355     & 10,539  & 1     & EM* \\ \hline
\end{tabular}
\end{table}

\subsection{Implementation Details}
\label{model}
% We train QBERT based on SBERT which is pre-trained with natural language inference dataset \cite{bowman2015large,williams2017broad} and semantic textual similarity dataset \cite{cer2017semeval}. 
For each input sequence, the length is limited to 35 tokens because we use two BERTs to read the sequence pair instead of concatenating two sequences into one as the input. Besides, most questions in the datasets have less than 35 tokens. The sequence is truncated at the end if it is longer than the limitation.

We train QBERT with the multiple negative ranking loss for QT and the online contrastive loss for QE. And we define the similarity threshold for QE based on the best accuracy on the training set. Then we evaluate the model on both QE classification and retrieval tasks. The QE retrieval candidate corpus is constructed by sampled queries in the QQP test set.

For QA, we train WikiQA with the online contractive loss and YQA and SQuAD with multiple negative ranking loss. This is because YQA and SQuAD only contain question answering pairs and do not come with negative samples. However, for WikiQA, there are questions with no answers in the dataset. Thus, a threshold is needed to identify if the closest candidate to the question is the high-confidence answer. The threshold is defined as the one that creates the best precision in the WikiQA training set.

% During inference, the model take the candidate sentence with the highest similarity as the answer. The candidate sentences are contained in WikiQA and SQuAD for each query.

The implementation of QBERT is based on PyTorch and SBERT. The margin for positive samples and negative samples is 0.5. We train the model for 5 epochs with a batch size of 32 and a learning rate of $2e-5$. 10\% of the training data is used for warm-up.

We train QBERT with one GeForce GTX TITAN X GPU. To train QBERT-OBO, it takes 45.5hr, and 93hr for QBERT-RR. Even though training the model is time-consuming, once trained, the model is much faster during inference. It takes 1.5ms, 5.44ms, 19.62ms, and 49.76ms per question in YQT, QQP, WikiQA, and SQuAD, respectively.

\section{Performance of QBERT} 
\label{performance}
We evaluate QBERT with YT for QT classification, QQP for QE classification and QE retrieval, and WikiQA and SQuAD for QA retrieval.

For QE, we evaluate classification and retrieval task accuracy with the QQP dataset. If the question pair has a similarity larger than the threshold, it is categorised as equivalent in classification. To perform similar question mining, we create a question corpus based on QQP. First, all the relevant questions for the given query are included in the dataset, ensuring that there is always a relevant question in the corpus. Second, we fill the rest of the corpus with irrelevant questions. There are 104,033 samples in total. While mining the similar questions from the corpus, the candidate with the highest similarity larger than the threshold is defined as the duplicate question.

We assess the QA performance only on WikiQA and SQuAD, because the answers in the YQA are paragraphs provided by Yahoo! users, and it is hard to construct a candidate corpus used for single-sentence answer retrieval. We count the number of questions that can correctly identify the answer (or \textit{None} for the questions without an answer) from the corpus while evaluating QA tasks. 

In WikiQA, the question is not guaranteed to have an answer. Therefore, for each question, the model takes the sentence with the highest cosine similarity score in the candidate set and compares it with the threshold. If the similarity is above the threshold and the sentence is labelled as a correct answer, then the prediction is correct. For SQuAD, each query has a corresponding answer in the given context. Thus, we take the sentence with the highest cosine similarity score as the candidate. Note that for SQuAD, the ground truth answer is a short answer phrase extracted from the given context. Since QBERT retrieves one sentence as the answer, we evaluate the exact match phrase in the sentence, which depends on whether the answer phrase is in the selected sentence.

To understand the performance of multi-task learning, we use SBERT, which is fine-tuned with natural language inference dataset \cite{bowman2015large,williams2017broad} and semantic textual similarity dataset \cite{cer2017semeval} as our baseline. We also compare QBERT with the single-tasks model. The result is shown in Table \ref{tab:result}. 

\begin{table*}[hbt]
\caption{The performance of QBERT-RR and QBERT-OBO compares with the performance of single-task SBERT trained on QT, QA and QE. SBERT without training on any question related dataset is used as the baseline. And we evaluate the QQP dataset on both classification and retrieval data.}
\label{tab:result}
\centering
\resizebox{\textwidth}{!}{
\begin{tabular}{l|l|l|l|l|l}
\multirow{2}{*}{\textbf{Curr.}} & \textbf{YQT}                           & \textbf{QQP-C}           & \textbf{QQP-R}                                                & \textbf{WikiQA}                                                      & \textbf{SQuAD}                         \\ \cline{2-6} 
                                & Acc.                                   & Acc.                                  & Acc./F1                                                              & Acc./F1                                                              & Acc.                                   \\ \hline \hline
\textbf{Baseline}                  & 35.27$\pm$0.58          & 74.80$\pm$0.32         & 54.53$\pm$0.89/53.01$\pm$0.82        & 77.46$\pm$2.82/58.24$\pm$12.48         & 67.04$\pm$0.93          \\
\textbf{QT}               & 72.44$\pm$0.39          &                                       &                                                                      &                                                                      &                                        \\
\textbf{QE}               &                                        & 89.79$\pm$0.23         & 56.98$\pm$0.65/55.36$\pm$0.60         &                                                                      &                                        \\
\textbf{QA}               &                                        &                                       &                                                                      & 79.05$\pm$5.89/72.50$\pm$11.08         & 78.59$\pm$0.82 \\
\textbf{OBO}              & 59.84$\pm$0.32          & 78.85$\pm$0.44         & 57.46$\pm$0.78/55.87$\pm$0.70          & 80.16$\pm$6.03/69.29$\pm$9.00          & 76.09$\pm$0.66          \\
\textbf{RR}               & 73.77$\pm$0.58 & 90.13$\pm$0.19 & 58.22$\pm$0.78/56.53$\pm$0.75 & 81.90$\pm$5.60/73.73$\pm$8.12 & 71.42$\pm$1.45         
\end{tabular}
}
\end{table*}

SBERT was only trained on natural language inference dataset and semantic textual similarity dataset containing sentence pairs with labels. It therefore, manages to detect similar question pairs albeit with poor performance. However, SBERT was not trained to group sentences with the same topic, and it is unable to identify the question topic. Since SBERT achieves a similar accuracy to other models on WikiQA dataset, it has a worse F1 score compared to others.

In Table \ref{tab:result}, model SBERT-QT, SBERT-QE, SBERT-QA represent single-task training. It leverages the same architecture as QBERT. However, for each task, it has a separate model. While fine-tuning the single-task model, we update both the BERT layer and task-specific layer for each dataset. The results show that QBERT-RR achieves a similar performance on most question datasets, except for retrieving answer from SQuAD, with the generalist representation.

The previous research \cite{xu2021makes} proved that the training curriculum is important for training a multi-task network. Thus, we investigate two different training strategies. The QBERT-OBO shows better performance on WikiQA; on the other hand, it has worse performance on QT and QE compared to the single-task models. When training QBERT-OBO, we train one dataset after another. This causes the model to ``forget'' what it learnt during the early stages.

\begin{figure}[hbt!]
    \centering
    \begin{subfigure}[]{0.45\textwidth}
        \centering
        \includegraphics[width=\textwidth]{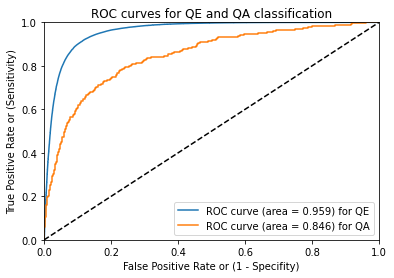}
        % \centering
        \caption{}
        \label{fig:roc}
    \end{subfigure}
        \begin{subfigure}[]{0.53\textwidth}
        \centering
        \includegraphics[width=\textwidth]{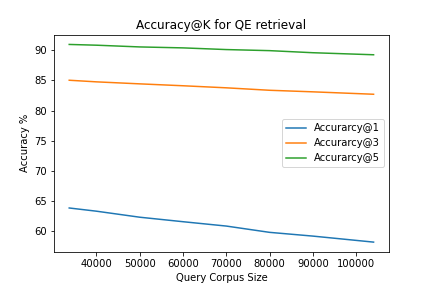}
        % \centering
        \caption{}
        \label{fig:accuracy@k}
    \end{subfigure}
    \caption{(a): ROC curve for QE and QA classification from model QBERT-RR. The black dashed line represents the performance of a random classifier. (b): Accuracy@K of different corpus sizes in QE retrieval task.}
\end{figure}

In contrast, while training with the RR curriculum, the model achieves a balanced performance on each task. Although QBERT-RR does not excel in any task compared to the single task model, it is able to generate a representation that can be used to perform a range of question tasks. Figure \ref{fig:roc} shows the performance of QE and QA classification tasks.

We also evaluate the accuracy@k among different retrieval corpus sizes for QE using the QQP test set. Accuracy@k is a top-k accuracy classification score. In QE, it counts the number of times where the relevant question is contained in the top k candidates. According to the results illustrated in figure \ref{fig:accuracy@k}, it is more challenging to retrieve among the larger corpus. When all the queries in the dataset are included in the retrieval corpus, the accuracy@1, accuracy@3, accuracy@5 are 58.24\%, 82.72\%, and 89.26\%, respectively. More than 80\% of the related questions are located in the top 3 candidates. However, only 58.150\% of them are the closest to the given query, which can be improved in the future.

Lastly, we notice one limitation while evaluating QA retrieval with the SQuAD dataset. When creating the candidate corpus, we leverage a sentence tokenizer to split the paragraph into sentences. However, the sentence tokenizer split the sentence based on the punctuation. For example, `` Washington, D.C.'' is considered two sentences: ``Washington, D.'' and ``C.''. During evaluation, we compare the selected sentence with the answer phrase. In this case, retrieving a sentence may yield an incomplete answer to a question.

\section{Conclusion}
In this paper, we propose a generalist model to process questions in a variety of tasks, namely Question Topic Classification, Equivalence Question Recognition, and Question Answering. The idea is that sometimes a generalist model can be useful even when it does not beat specialist models at their own speciality.

We fine-tune SBERT as a generalist model for processing questions. We observe that one version of the generalist model QBERT-RR turns out to perform similar to the specialists in many cases except for QA retrieval on the SQuAD dataset. The specialist methods used here for comparison are SBERT models fine-tuned respectively on QT, QE (classification data) and QA (both datasets). Instead, another generalist method QBERT-OBO performs worse than the specialists on QT and QE (classification). The reasons for this performance need to be further investigated, but it may happen because the OBO curriculum results in forgetting the tasks that are learnt in the earlier training stage.

In the future, it would also be useful to experiment with more tasks that can be represented with sentence embedding.

\bibliographystyle{unsrt}  
\bibliography{references}  %%% Remove comment to use the external .bib file (using bibtex).


% George Kour and Raid Saabne.
% \newblock Real-time segmentation of on-line handwritten arabic script.
% \newblock In {\em Frontiers in Handwriting Recognition (ICFHR), 2014 14th
%   International Conference on}, pages 417--422. IEEE, 2014.

% George Kour and Raid Saabne.
% \newblock Fast classification of handwritten on-line arabic characters.
% \newblock In {\em Soft Computing and Pattern Recognition (SoCPaR), 2014 6th
%   International Conference of}, pages 312--318. IEEE, 2014.

% \bibitem{hadash2018estimate}
% Guy Hadash, Einat Kermany, Boaz Carmeli, Ofer Lavi, George Kour, and Alon
% \newblock Estimate and replace: A novel approach to integrating deep neural
%   networks with existing applications.
% \newblock {\em arXiv preprint arXiv:1804.09028}, 2018.

% \end{thebibliography}

\end{document}